\documentclass[conference]{IEEEtran}
\IEEEoverridecommandlockouts
\usepackage{amsmath,amssymb,amsfonts}
\usepackage{algorithmic}
\usepackage{graphicx}
\usepackage{textcomp}
\usepackage{xcolor}
\usepackage[breaklinks]{hyperref}

\usepackage[backend=bibtex]{biblatex}
\usepackage{tabularray}
\def\BibTeX{{\rm B\kern-.05em{\sc i\kern-.025em b}\kern-.08em
    T\kern-.1667em\lower.7ex\hbox{E}\kern-.125emX}}
\begin{document}

\title{CCSPNet-Joint: Efficient Joint Training Method for Traffic Sign Detection Under Extreme Conditions\\

\thanks{Our code is at \href{https://github.com/HaoqinHong/CCSPNet-Joint}{https://github.com/HaoqinHong/CCSPNet-Joint}.}
}

\author{\IEEEauthorblockN{1\textsuperscript{st} Haoqin Hong}
\IEEEauthorblockA{\textit{Institute of Innovation \& Entrepreneurship, Hanhong College} \\
\textit{Southwest University}\\
Chongqing, China \\
honghaoqin@emial.swu.edu.cn}
\and
\IEEEauthorblockN{2\textsuperscript{nd} Yue Zhou}
\IEEEauthorblockA{\textit{School of Artificial Intelligence} \\
\textit{Southwest University}\\
Chongqing, China \\
zhouyuenju@163.com}
\and
\IEEEauthorblockN{3\textsuperscript{rd} Xiangyu Shu}
\IEEEauthorblockA{\textit{College of Computer and Information Science} \\
\textit{Southwest University}\\
Chongqing, China \\
shuxy6263@163.com}
\and
\IEEEauthorblockN{4\textsuperscript{th} Xiaofang Hu}
\IEEEauthorblockA{\textit{School of Artificial Intelligence} \\
\textit{Southwest University}\\
Chongqing, China \\
huxf@swu.edu.cn}
}

\maketitle

\begin{abstract}
Traffic sign detection is an important research direction in intelligent driving. Unfortunately, existing methods often overlook training methods specifically designed for extreme conditions such as fog, rain, and motion blur. Moreover, the end-to-end training strategy for image denoising and object detection models fails to utilize inter-model information effectively. To address these issues, we propose \textbf{CCSPNet}, an efficient feature extraction module based on Contextual Transformer and CNN, capable of effectively utilizing the static and dynamic features of images, achieving faster inference speed and providing stronger feature enhancement capabilities. Furthermore, we establish the correlation between object detection and image denoising tasks and propose a joint training model, \textbf{CCSPNet-Joint}, to improve data efficiency and generalization. Finally, to validate our approach, we create the \textbf{CCTSDB-AUG} dataset for traffic sign detection in extreme scenarios. Extensive experiments have shown that CCSPNet achieves state-of-the-art performance in traffic sign detection under extreme conditions. Compared to end-to-end methods, CCSPNet-Joint achieves a 5.32\% improvement in precision and an 18.09\% improvement in mAP@.5.\par
\end{abstract}


\section{Introduction}
Traffic sign detection (TSD) plays a significant role in the field of intelligent driving by providing vital road information to intelligent driving systems, enabling accurate recognition for subsequent decision-making processes\cite{faisal2019understanding, yang2015towards}. Traffic sign detection algorithms utilize computer vision techniques to rapidly and accurately identify and extract information from images or video data pertaining to traffic signs\cite{saadna2017overview}. The application of such algorithms aids intelligent vehicles in the real-time acquisition of road sign information, enhancing driving safety and overall driving efficiency.\par

Early traffic sign detection algorithms primarily relied on traditional computer vision techniques, such as image edge detection \cite{maini2009study}, image filtering algorithms \cite{he2012guided}, and morphological image processing \cite{dougherty2003hands}. With the advancement of deep learning technology\cite{lecun1998gradient}, deep learning-based traffic sign detection algorithms have gained widespread adoption. Applying deep learning methods to autonomous driving traffic sign detection improves detection accuracy, speed, and enhances the model's generalization capability. Convolutional neural networks (CNNs) demonstrate strong advantages in traffic sign detection. Compared to other algorithms, CNNs extract more comprehensive features, effectively improving recognition accuracy and speed. CNNs process images through convolution and pooling operations, extracting different features, and automatically learning optimal features, thereby enhancing the accuracy of traffic sign detection. Additionally, CNNs exhibit adaptability, allowing for adjustments in network depth and structure to accommodate different traffic sign detection tasks\cite{krizhevsky2012imagenet}. However, CNN-based methods alone may not effectively capture global information in complex traffic scenarios. Therefore, it is crucial to design an effective model that can efficiently extract local feature information while improving the capture of global information in traffic scenes.\par

The backbone network based on Transformer \cite{vaswani2017attention} has better capabilities in capturing global information compared to the backbone network based on CNNs. Transformer has been widely used in various visual tasks, including image classification \cite{dosovitskiy2020image, wang2021pyramid}, object detection \cite{carion2020end}, and semantic segmentation\cite{liu2021swin, liu2022swin}, among others, for feature extraction. The advantage of Transformer lies in its ability to handle long-range dependencies and parallel computation, effectively capturing crucial information in input sequences. It enables end-to-end training and inference for different tasks. However, existing traffic sign detection models have not fully utilized the advantages of Transformer. In practical traffic scenarios, extracting global features allows for comprehensive consideration of overall shape, color, and other information related to traffic signs. These features possess certain scale and rotation invariance, thereby enhancing the model's generalization capability. Such comprehensive feature representations contribute to a better understanding and handling of traffic sign detection tasks at different scales and rotation angles. In traffic sign detection, both local and global features are important, as local features aid in detecting specific sign details and shapes, while global features provide contextual information for accurate localization and recognition of sign positions and categories. Therefore, combining the advantages of convolutional neural networks and Transformer models, We have designed an efficient feature extraction module called Cross Contextual Stage Partical Network (CCSPNet). It effectively captures both local and global information of features and exhibits real-time capability, enabling efficient traffic sign detection tasks.\par

Compared to traffic signs in other countries, Chinese traffic signs have unique characteristics in terms of colour, patterns, fonts, reflectivity, and other aspects. Analysis of the existing datasets reveals the presence of extreme conditions in traffic scenes, such as rainy weather, foggy conditions, and motion blur. These complex scenarios significantly impact the perception of traffic signs by intelligent vehicles, making it challenging for intelligent driving systems to accurately acquire target information and affecting the decision-making process. Therefore, existing traffic sign detection tasks face two main challenges: designing a traffic sign detection algorithm that ensures both model accuracy and real-time performance, and developing an adaptive denoising algorithm to improve the performance of object detection effectively. Based on the above, the main contributions of this article are as follows:

\begin{itemize}

    \item We have designed an efficient feature extraction module named Cross Contextual Stage Particle Network (CCSPNet) based on Contextual Transformer and CNN. It exhibits enhanced sensitivity to dynamic features in images, thereby possessing stronger visual representation capabilities and effectively capturing both local and global feature information. Subsequently, we introduced the EfficientViT backbone network and proposed the object detection model YOLO-CCSPNet, which achieves higher detection accuracy.

    \item We developed a joint training method named CCSPNet-Joint that establishes a correlation between image denoising and object detection. This approach effectively addresses the challenges of traffic sign detection in extreme conditions such as fog, rain, and motion blur.
    
    \item We simulated the CCTSDB dataset using image augmentation techniques to generate the CCTSDB-AUG dataset, which consists of traffic sign images under extreme conditions. Extensive experiments have been conducted, demonstrating that both our model and training method have achieved the best performance.

\end{itemize}

\section{Related Work}
\subsection{Traffic sign detection} 
Traffic sign detection refers to using computer vision techniques to detect and recognize various traffic signs in road traffic scenes, such as directional signs, prohibition signs, warning signs, etc. There are several existing methods for object detection, primarily divided into CNN-based \cite{krizhevsky2012imagenet} and Transformer-based \cite{vaswani2017attention} object detection models. CNN-based object detection models, such as Faster R-CNN \cite{ren2015faster} and YOLO\cite{bochkovskiy2020yolov4, redmon2016you, jocher2022ultralytics, wang2023yolov7}, have a long history of research. The advantage of CNNs lies in their ability to learn how to extract features from images through training and can be fine-tuned for different tasks. Additionally, CNNs can handle inputs of various sizes and detect multiple objects. However, CNNs may not effectively capture global information, leading to the utilization of Transformer models in object detection research. Transformer models introduce the self-attention mechanism, which enables better capturing of long-range dependencies in sequences and enhances the expressive power of the model. Furthermore, the multi-head attention mechanism in Transformer models allows attention to be distributed across different spatial positions, resulting in stronger modelling capabilities. Moreover, Transformer models can be computed in parallel, leading to faster computation speeds. Notable Transformer-based object detection models include DETR\cite{carion2020end}, Swin Transformer\cite{liu2021swin}\cite{liu2022swin}, among others. These methods effectively address the limitations of CNNs in handling objects of different scales, capturing global information, and modelling long-range dependencies. Many traffic sign detection models have been proposed using the above-mentioned methods: Y. Lu et al. \cite{lu2018traffic} proposed an attention-based approach for traffic signal detection and classification. This method utilizes attention mechanisms to extract features from traffic signals and employs convolutional neural networks for classification. Y. Dong et al. \cite{dong2021accurate} introduced a traffic sign detection and recognition method based on Swin Transformer, incorporating cross-stage feature fusion and adaptive mask pooling. This method further enhances detection performance. M. Hou et al. \cite{hou2022traffic} presented a traffic sign detection and recognition system based on lightweight multitask learning. The system utilizes a multitask learning framework for simultaneous detection and recognition, enhancing feature representation through attention mechanisms.\par

\subsection{Image denosing} 
Traffic sign image denoising primarily involves techniques such as rain removal, fog removal, and motion deblurring, which aim to enhance the image quality in complex traffic scenes. Rain removal refers to the process of eliminating image blur and distortion caused by raindrop occlusions\cite{yasarla2020syn2real}\cite{guo2021efficientderain}. Fog removal is aimed at reducing the haze and atmospheric scattering effects in images\cite{he2010single}\cite{zheng2021ultra}. Motion deblurring is the process of recovering images affected by non-uniform blurring caused by camera shake, scene depth variations, and the motion of multiple objects\cite{tsai2022stripformer}. \par

\subsection{Joint Training} 
Current existing joint training methods include end-to-end joint training and joint transfer learning methods, which enable the model to simultaneously learn image denoising and object detection tasks by training different parts of the model together. K. Chen et al. \cite{chen2019hybrid} proposed a multi-stage cascaded model called Hybrid Task Cascade (HTC), which integrates three tasks: bounding box detection, semantic segmentation, and instance segmentation, into a unified framework. They utilize multi-task learning to enhance the complementarity and collaboration between the tasks. Y. Xu et al. \cite{xu2023multi} introduced a new multi-task learning method that applies knowledge distillation in object detection. This method improves the model's performance by distilling knowledge between different tasks.\par

Existing methods typically treat image denoising and traffic sign detection as two separate modules that are run end-to-end. However, we aim to design a model that can better utilize the information between the image-denoising task and the traffic sign detection task to improve accuracy and speed in complex real-world scenarios. Joint training provides an excellent solution to this problem. Joint training of image denoising and object detection is a method that combines the training of both tasks to train a model, thus enabling the model to learn both image denoising and object detection tasks simultaneously by training different parts of the model together.\par

\section{METHODOLOGY}
\subsection{Outline}

\begin{figure}[t]
  \centering
  \centerline{\includegraphics[width=8.5cm]{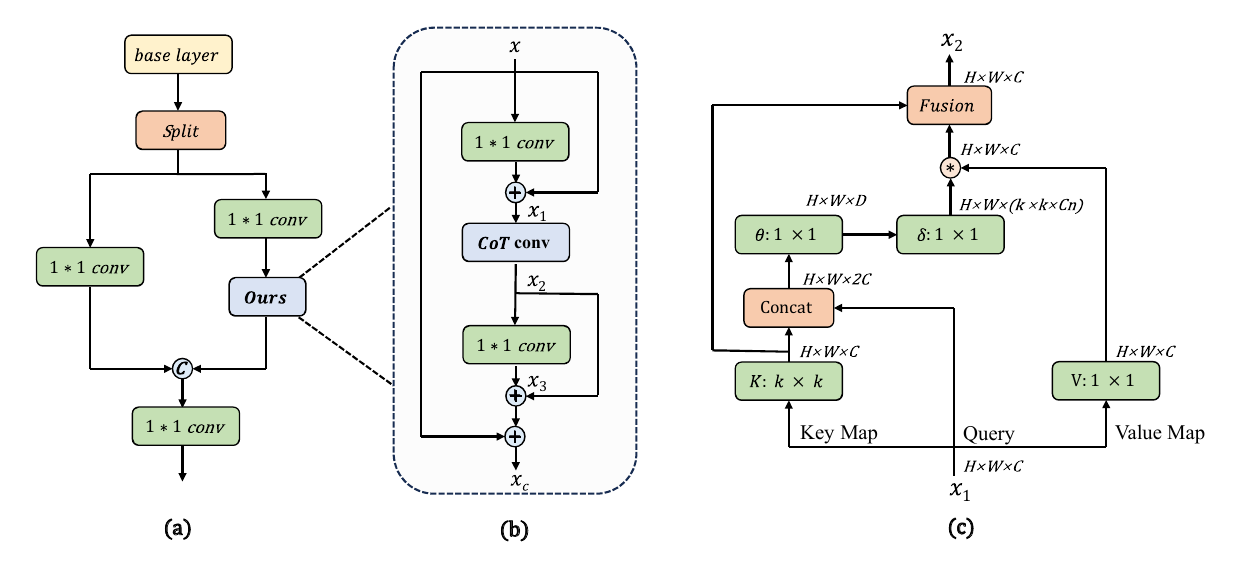}}
   \caption{The feature enhancement module that we proposed:
(a) Architecture of our feature enhancement module.
(b) The structure of the CCSPNet.
(c) The structure of the CoT.}
\label{fig:res}
\end{figure}

First, we introduced the concept of Contextual Transformer to design an effective feature extraction module called CCSPNet. This module not only retains the local feature extraction capability of CNNs but also enhances the model's global feature extraction ability, making it more sensitive to dynamic contextual features in images and strengthening visual representation capability. Second, to fully leverage the global feature extraction capability of Transformers, we optimized the backbone network by introducing EfficientViT, a backbone network that balances speed and accuracy. EfficientViT effectively improves the model's accuracy while maintaining computational efficiency. Finally, to further enhance the model's detection accuracy in extreme conditions, we combined an image-denoising model and proposed a joint training approach called CCSPNet-Joint. This approach further improves the model's generalization for complex traffic scenes. 

\subsection{Cross Contextual Stage Partical Network}
CNNs have been commonly used in object detection models for their strong local feature extraction\cite{ren2015faster}. Deep networks enable complex recognition tasks. To overcome CNNs' limitations in global feature extraction, Transformers have been integrated into object detection models, improving their generalization ability\cite{carion2020end, liu2021swin}. In this paper, we have improved the feature enhancement part of YOLOv5 by effectively combining the strengths of CNNs and Transformers\cite{wang2020cspnet}. We have designed a new module called Cross Contextual Stage Partial Network (CCSPNet), and its model structure is depicted in \textbf{Fig. 1(b)}.\par

For the input feature map $X$, it undergoes a $1 \times 1$ convolutional layer with a residual structure to obtain the feature map $X_1$. After the initial feature transformation, we utilize the Contextual Transformer(CoT)\cite{li2022contextual} structure to perform global feature extraction on $X_1$. The CoT is an improved Transformer structure, and its model structure is depicted in \textbf{Fig. 1(c)}.\par

The CoT layer performs a $k \times k$ grouped convolution operation on the input feature map, capturing local information of the image. Then, the local information is concatenated with the original information, followed by consecutive convolutional layers for further feature map computation. The $SoftMax$ operation is applied to compute self-attention with the Value feature map, obtaining the global information of the image. \par

First, the input feature map $X_{1}$ is transformed into query $Q = X_{1}$, key $K = X_{1}$, and value $V = X_{1}W_{v}$. Then, a $3 \times 3$ convolution operation is applied to $K$, resulting in the static contextual feature $K_{1}$. $K_{1}$ is concatenated with $Q$, and passed through two $1 \times 1$ convolutions to obtain the attention matrix $A$. Next, the attention matrix $A$ is used to aggregate the values $V$ through weighted summation, resulting in the dynamic contextual feature $K_{2}$. Finally, $K_{1}$ and $K_{2}$ are fused together to form the output $X_{2}$. The calculation formula is as follows:\par

\begin{align}
    K_{1} = K \cdot W_{k} \label{eq:1} \\
    A = [K_{1}, Q] \cdot W_{\theta} \cdot W_{\delta} \label{eq:2} \\
    K_{2} = V \ast A \label{eq:3} \\ 
    X_{2} = K_{1} + K_{2} \label{eq:4}
\end{align}

In this process, $W_{v}$, $W_{\theta}$, and $W_{\delta}$ are $1 \times 1$ convolutional matrices, and $W_{k}$ is a $3 \times 3$ convolutional matrix. The $\ast$ represents the local matrix multiplication operation.\par

The CoT layer outputs $X_{2}$, which is then subjected to another convolutional computation with a residual structure to obtain the feature map $X_{3}$. Finally, the input feature map $X$ is fused with $X_{3}$ to obtain the feature map $X_{c}$, which contains rich local features and global information. The calculation formula is as follows:\par

\begin{equation}
    X_{c} = X + X_{3} \label{eq}
\end{equation}

The architecture for the final feature enhancement is illustrated in \textbf{Fig. 1(a)}. We replaced the convolutional recurrent layers in the original YOLOv5 feature enhancement module. This design effectively utilizes contextual information between input keys to guide the learning of the dynamic attention matrix, thereby enhancing visual representation capability and improving the model's detection accuracy.\par

\subsection{Transformer-based backbone network}

\begin{figure}[htbp]
  \centering
  \centerline{\includegraphics[width=8.5cm]{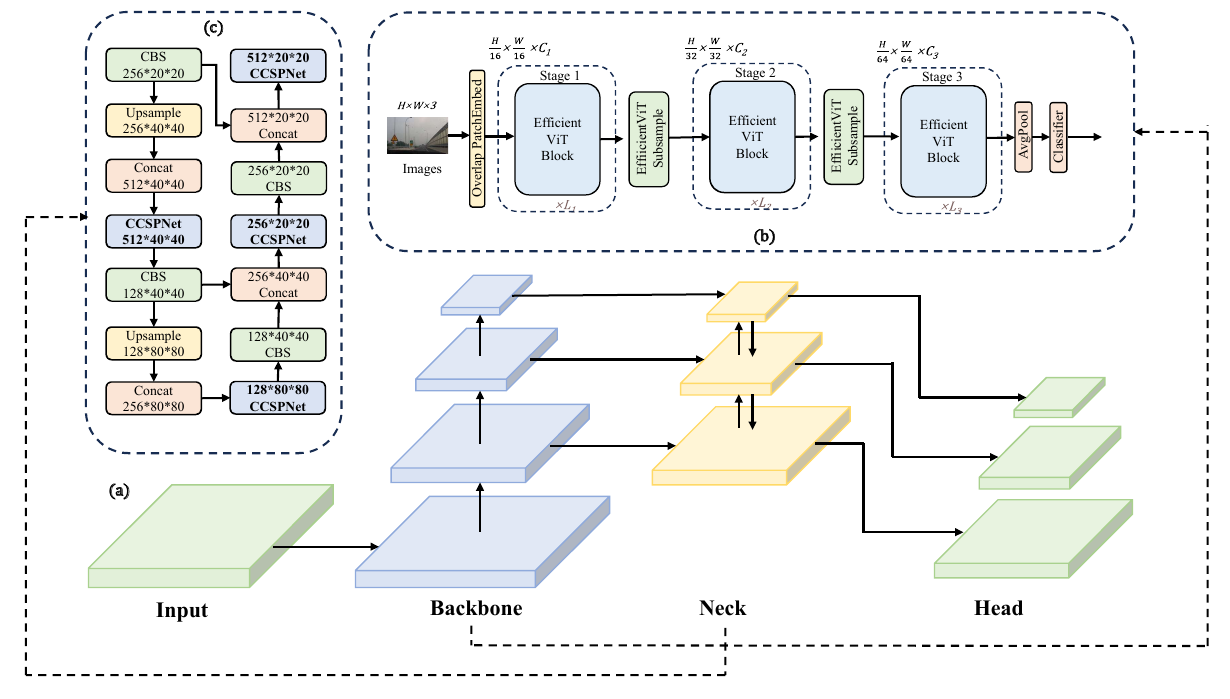}}
   \caption{The object detection model YOLO-CCSPNet in this article: (a) a framework for one-stage object detection, (b) a backbone network based on EfficientViT, and (c) a Neck module based on CCSPNet.}
\label{fig:res}
\end{figure}

\begin{figure*}[htbp]
  \centering
  \centerline{\includegraphics[width=14cm]{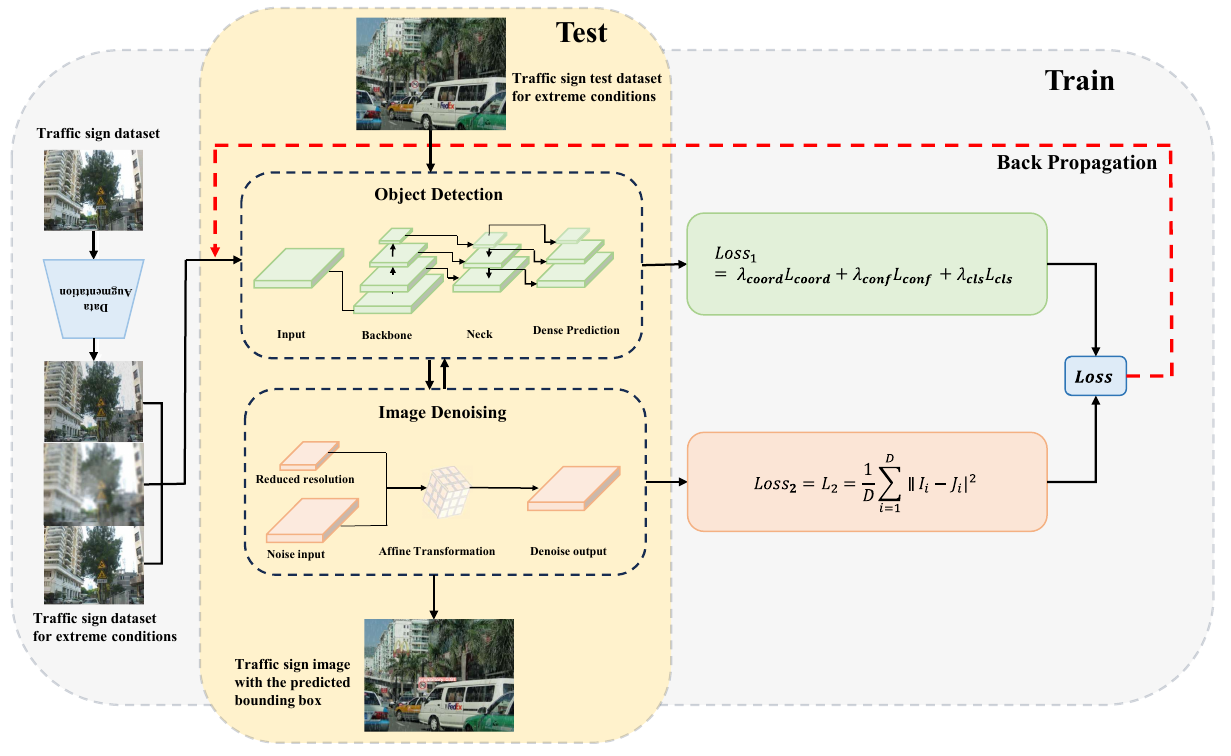}}
  \caption{The proposed joint training method CCSPNet-Joint for traffic sign detection in extreme conditions.}
  \label{fig:res}
\end{figure*}

We choose EfficientViT\cite{liu2023efficientvit} as the backbone network for efficient traffic sign detection, prioritizing its faster inference speed and superior feature extraction capabilities over Swin Transformer\cite{liu2021swin}. EfficientViT improves memory and parameter efficiency, reduces computational costs, and maintains high performance. By adopting EfficientViT, we aim to reduce deployment costs and difficulties associated with Vision Transformers while leveraging their high model capacity. EfficientViT enables real-time and efficient inference for practical traffic sign detection scenarios. Therefore, the object detection model we proposed, YOLO-CCSPNet, has a network architecture as shown in \textbf{Fig. 2(a)}.\par


\subsection{Joint Training Method}
In real-world scenarios, foggy, rainy, and blurry conditions are common occurrences. A common approach is applying denoising techniques to the images to enhance their visual clarity\cite{tsai2022stripformer, zheng2021ultra, yasarla2020syn2real, guo2021efficientderain, he2010single}. However, denoising inevitably alters the original image features, which can affect the ability of object detection models to interpret the feature maps.
Therefore, in order to enhance the detection accuracy of models in extreme weather conditions, we propose joint training with a denoising model. This approach combines the denoising model with the object detection model, allowing them to learn and adapt together. By integrating the denoising process into the training pipeline, the model can better handle noisy and challenging weather conditions, ultimately improving the overall performance of the object detection task.\par


We have chosen the 4kDehazing model\cite{zheng2021ultra} as our baseline for image denoising. It is an ultra-high-definition image-denoising method that utilizes multi-guided bilateral learning. This approach employs a deep convolutional neural network to construct an affine bilateral grid, preserving detailed edges and textures in the image. The network consists of three deep CNNs: the first CNN extracts haze-relevant features at a reduced resolution of the hazy input and fits locally affine models in the bilateral space; the second CNN is used to learn multiple full-resolution guidance maps corresponding to the learned bilateral model, enabling the reconstruction of high-frequency feature maps through multi-guided bilateral upsampling; finally, the third CNN merges the high-quality feature maps into a dazed image.\par

The specific process of joint training is as shown in \textbf{Fig. 3}. To establish a better connection between the image-denoising model and the object detection model, while effectively addressing overfitting during model training, we propose a loss function to constrain the training process of the object detection model. Our loss function consists of two parts, $\mathcal{L}_1$ represents the loss function for object detection, which is identical to the loss function used in YOLOv4\cite{bochkovskiy2020yolov4}, $\mathcal{L}_2$ represents the loss function for image denoising, where we adopt the same loss function as in 4kDehazing\cite{zheng2021ultra}.\par

We use the cross-entropy loss function to calculate the classification loss of target categories. Assuming there are $S$ cells, each cell has $B$ anchor boxes, and each anchor box corresponds to $C$ categories. For each cell $i$ and anchor box $j$, we define an indicator function $\mathbb{1}_{ij}^{\text{obj}}$, which indicates whether the anchor box is responsible for detecting the target. The classification loss calculation for each cell $i$ and anchor box $j$ is as follows:\par

\begin{equation}
    \mathcal{L}_{\text{cls}_{ij}} = \sum_{i=0}^{S^2}\sum_{j=0}^{B}\mathbb{1}_{ij}^{\text{obj}}\sum_{c=0}^{C}(p_{ij}(c)-\hat{p}_{ij}(c))^2 \label{eq}
\end{equation}

Where $p_ij(c)$ represents the predicted class probability, $\hat{p}_{ij}(c)$ represents the true class probability. In this case, the squared error is used as the metric for classification loss.\par

We use the Mean Squared Error (MSE) loss function to calculate the localization loss of bounding boxes. For each cell $i$ and anchor box $j$, we define an indicator function $\mathbb{1}_{ij}^{\text{obj}}$, which indicates whether the anchor box is responsible for detecting the target. The formula for calculating the localization loss for each cell $i$ and anchor box $j$ is as follows:

\begin{equation}
\begin{aligned}
    \mathcal{L}_{\text{loc}_{ij}} = \sum_{i=0}^{S^2}\sum_{j=0}^{B}\mathbb{1}_{ij}^{\text{obj}}[(x_{ij}-\hat{x}_{ij})^2+(y_{ij}-\hat{y}_{ij})^2] + &\\ \sum_{i=0}^{S^2}\sum_{j=0}^{B}\mathbb{1}_{ij}^{\text{obj}}[(\sqrt{w_{ij}}-\sqrt{\hat{w}_{ij}})^2 + (\sqrt{h_{ij}}- & \sqrt{\hat{h}_{ij}})^2]
\end{aligned}
\end{equation}

Where $x_ij$ and $y_ij$ are the predicted coordinates of the bounding box's center, $w_ij$ and $h_ij$ are the predicted width and height of the bounding box.
$\hat{x}_{ij}$, $\hat{y}_{ij}$, $\hat{w}_{ij}$, and $\hat{h}_{ij}$ are the true coordinates, width, and height of the bounding box, respectively. In this case, the squared error is used as the metric for the localization loss.\par

We use the binary cross-entropy loss function to calculate the loss for object confidence. For each cell $i$ and anchor box $j$, two indicator functions are defined: $\mathbb{1}_{ij}^{\text{obj}}$ and $\mathbb{1}_{ij}^{\text{noobj}}$, which indicate whether the anchor box is responsible for detecting the target or not detecting the target, respectively. The formula for calculating the object confidence loss for each cell $i$ and anchor box $j$ is as follows:\par

\begin{equation}
\begin{aligned}
    \mathcal{L}_{\text{obj}_{ij}} = &\sum_{i=0}^{S^2}\sum_{j=0}^{B}\mathbb{1}_{ij}^{\text{obj}}\left[(1 - p_{ij}^{\text{obj}})^2\right] \\
    &+ \lambda_{\text{noobj}}\sum_{i=0}^{S^2}\sum_{j=0}^{B}\mathbb{1}_{ij}^{\text{noobj}}\left[(0 - p_{ij}^{\text{obj}})^2\right]
\end{aligned}
\end{equation}

Where $p_{ij}^{\text{obj}}$ represents the predicted object confidence score, $\mathbb{1}{ij}^{obj}$ and $\mathbb{1}{ij}^{noobj}$ are indicator functions, indicating whether the anchor box is responsible for detecting the target or not, $\lambda_{\text{noobj}}$ is a weight used to balance the loss for responsible and non-responsible anchor boxes. The object confidence loss uses the squared error as the metric. Finally, the total loss function in the object detection module is obtained by summing the weighted losses of these three components:\par

\begin{equation}
    \mathcal{L}_1 = \mathcal{L}_{\text{total}} = \lambda_{\text{cls}}\mathcal{L}_{\text{cls}} + \lambda_{\text{loc}}\mathcal{L}_{\text{loc}} + \lambda_{\text{obj}}\mathcal{L}_{\text{obj}} 
\end{equation}

Where $\lambda_{\text{cls}}$, $\lambda_{\text{loc}}$ and $\lambda_{\text{obj}}$ are the weights used to balance different loss terms.\par

4KDhazing optimizes the weights and biases of the proposed network by minimizing the $\mathcal{L}_2$ loss on the training set, the $\mathcal{L}_2$ loss is sufficient to generate denoised results with clear and vivid colours, its calculation formula is as follows:\par

\begin{equation}
    \mathcal{L}_2 = \frac{1}{D}\sum_{i=1}^{D}{\|I_i - J_i|^2} \label{eq}
\end{equation}

Where $D$ is the number of images in the training dataset, $I_i$ is the denoised image result predicted by the network, and $J_i$ is the corresponding Ground Truth image.\par

The overall computation formula for the joint training loss is given below:\par

\begin{equation}
    \mathcal{L}_{joint} = \alpha * \mathcal{L}_1 + \beta * \mathcal{L}_2 \label{eq}
\end{equation}

Where $\alpha$ and $\beta$ are hyperparameters that control the relative importance of $\mathcal{L}_1$ and $\mathcal{L}_2$, respectively. During the experiments, we observed that the joint training model achieved favourable training results when setting $\alpha$ = $\beta$ = 0.5. The joint training loss function optimizes the training process of the object detection model through backpropagation. Establishing a connection between the loss functions of the object detection module and the image denoising module aims to ensure convergence of the entire global model. This ensures that all components of the model converge together during training.\par

\section{Experimental Results}
\subsection{Dataset}

\begin{figure}[htbp]
  \centering
  \centerline{\includegraphics[width=8.5cm]{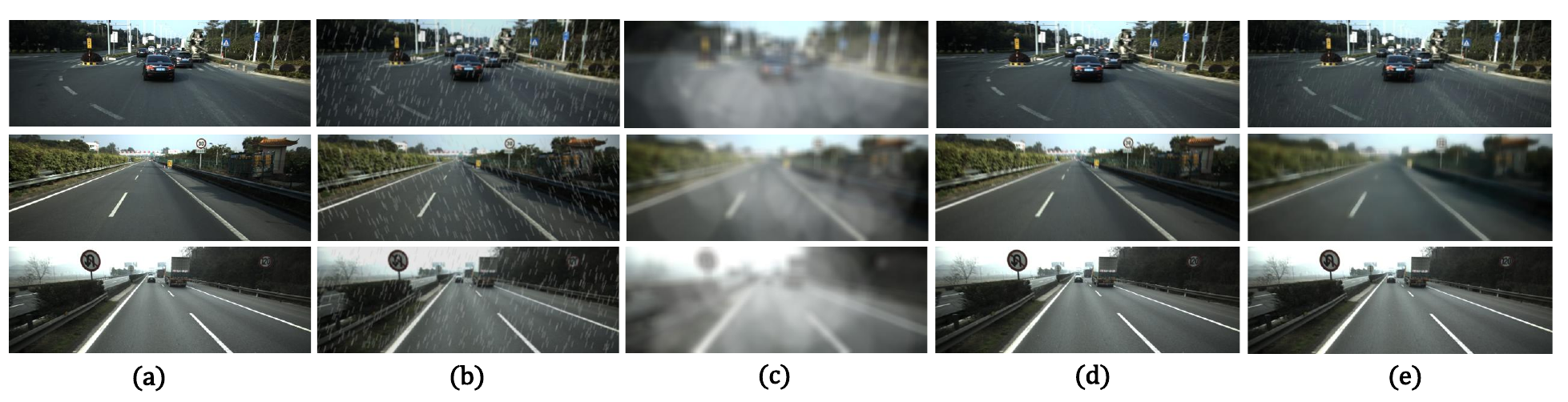}}
\caption{Experimental dataset: (a) Original images from CCTSDB, (b-d) Augmented images from CCTSDB-AUG with rain, fog, and motion blur, (e) Images processed by 4kDehazing in CCTSDB-AUG for rain removal, fog removal, and motion blur removal, arranged from top to bottom.}
\label{fig:res}
\end{figure}

\begin{table*}[h]
\caption{Comparison of Traffic Sign Detection Models}
\centering
\resizebox{12.5cm}{!}
{
\begin{tblr}{
  cells = {c},
  hline{1,12} = {-}{0.08em},
  hline{2} = {-}{},
    hline{4} = {-}{},
}
Model                                                     & Dataset    & Parameters & GFLOPs & Precision & Recall & map@.5 & map@.75 & FPS  \\
YOLOv5\textbf{(baseline)}\cite{jocher2022ultralytics}     & CCTSDB     & 46.12M     & 107.7  & 0.968     & 0.914  & 0.952  & 0.660   & 69.3 \\
YOLO-CCSPNet\textbf{(Ours)}                                           & CCTSDB     & 36.68M     & 65.4   & 0.974     & 0.924  & 0.958  & 0.664   & 46.3 \\
Faster RCNN\cite{ren2015faster}                           & CCTSDB-AUG & 41.13M     & 193.79 & 0.274     & 0.358  & 0.51   & 0.267   & 44.1 \\
DETR\cite{carion2020end}                                  & CCTSDB-AUG & 41.28M     & 91.62  & 0.438     & 0.648  & 0.858  & 0.388   & 41.6 \\
Swin Transformer\cite{liu2022swin}\cite{he2017mask}       & CCTSDB-AUG & 47.38M     & 145.87 & 0.470     & 0.535  & 0.768  & 0.534   & 30.9 \\
YOLOv4\cite{bochkovskiy2020yolov4}                        & CCTSDB-AUG & 63.94M     & 141.93 & 0.903     & 0.732  & 0.817  & 0.536   & 9.9  \\
YOLOv7\cite{wang2023yolov7}                               & CCTSDB-AUG & 36.49M     & 103.2  & 0.801     & 0.669  & 0.734  & 0.441   & 14.2 \\
YOLOv8l\cite{terven2023comprehensive}             & CCTSDB-AUG & 43.72M     & 162.7  & 0.902     & 0.771  & 0.870  & 0.518   & 98.2  \\
YOLOv5l\textbf{(baseline)}\cite{jocher2022ultralytics}    & CCTSDB-AUG & 46.12M     & 107.7  & 0.914     & 0.752  & 0.838  & 0.511   & 65.8 \\
YOLO-CCSPNet\textbf{(Ours)}                                           & CCTSDB-AUG & 36.68M     & 65.4   & \textbf{0.917}     & \textbf{0.774}  & \textcolor{red}{0.861} $\uparrow$ \textcolor{cyan}{+0.023}  & \textcolor{red}{0.518} $\uparrow$ \textcolor{cyan}{+0.007} & 44.4   
\end{tblr}
}
\end{table*}

The CSUST Chinese Traffic Sign Detection Benchmark (CCTSDB)\cite{zhang2017real, zhang2020lightweight} is an existing dataset for traffic sign detection. It consists of nearly 20,000 images of Chinese road traffic scenes, including around 40,000 annotated images of traffic signs. While most scenes in the dataset are captured under natural weather conditions, challenges include foggy, rainy, and blurry perspectives. To facilitate our research, we created a dataset called CCTSDB-AUG based on CCTSDB. This augmented dataset includes images with foggy, rainy, and blurry perspectives. We applied random haze, raindrop, and motion blur effects to generate these augmented images, simulating real-world extreme conditions. Image augmentation was performed using the Albumentations library in Python, allowing us to construct images with various weather effects. The CCTSDB-AUG dataset contains images with different extreme conditions, as illustrated in \textbf{Fig. 4(b-d)}. These extreme conditions are proportionally represented throughout the dataset.\par

\subsection{Experimental Settings}
The model construction in our work is implemented using the PyTorch framework and the experiments are conducted on a single NVIDIA GeForce RTX 4080 Laptop GPU. We apply the same data augmentation strategy to all the comparative models to control variables and ensure fair comparisons. The data augmentation strategy consists of only random horizontal flipping. Additionally, all experiments utilize the same input size and scaling strategy.\par

\subsection{Comparison Experiment}
Firstly, we selected seven mainstream convolutional-based or Transformer-based object detection models for comparison. The experimental results are shown in \textbf{Table \uppercase\expandafter{\romannumeral1}}. Through comparative experiments, it can be observed that our proposed model CCSPNet achieves improved detection accuracy compared to the baseline model YOLOv5, with an increase of 2.74\% in mAP@.50 and 1.37\% in mAP@.75. The final precision reaches 0.917, and the frames per second is 44.4. In comparison to two-stage and transformer-based object detection models, our model demonstrates comparable detection accuracy and faster inference speed, meeting real-time requirements. Notably, our model achieves the highest precision among all the compared models.\par


\begin{table}[h]
\centering
\caption{Comparison Experiments between Direct Training, End-to-End Training, and Joint Training}
\begin{tabular}{lllll}
\hline
\multicolumn{1}{c|}{\textbf{Model}}    & \multicolumn{1}{c}{\textbf{Precision}} & \multicolumn{1}{c}{\textbf{Recall}} & \multicolumn{1}{c}{\textbf{map@.5}} & \multicolumn{1}{c}{\textbf{map@.75}} \\ \hline
\multicolumn{5}{l}{\textbf{Direct-Training Method}}                                                                                                                                                \\ \hline
\multicolumn{1}{l|}{YOLOv5l(baseline)} & 0.914                                  & 0.752                               & 0.838                               & 0.511                                \\
\multicolumn{1}{l|}{YOLO-CCSPNet}      & 0.917                                  & 0.774                               & 0.861                               & 0.518                                \\ \hline
\multicolumn{5}{l}{\textbf{End-To-End Training Method}}                                                                                                                                            \\ \hline
\multicolumn{1}{l|}{YOLOv5l(baseline)} & 0.901                                  & 0.648                               & 0.727                               & 0.463                                \\
\multicolumn{1}{l|}{YOLO-CCSPNet}      & 0.903                                  & 0.704                               & 0.774                               & 0.467                                \\ \hline
\multicolumn{5}{l}{\textbf{Joint-Training Method}}                                                                                                                                                 \\ \hline
\multicolumn{1}{l|}{YOLOv5l(baseline)} & 0.945                                  & 0.819                               & 0.879                               & 0.553                                \\
\multicolumn{1}{l|}{CCSPNet-Joint}      & \textbf{0.951}                                  & \textbf{0.842}                               & \textbf{0.914}                               & \textbf{0.578}                                \\ \hline
\end{tabular}
\end{table}


\begin{figure}[h]
  \centering
  \centerline{\includegraphics[width=8cm]{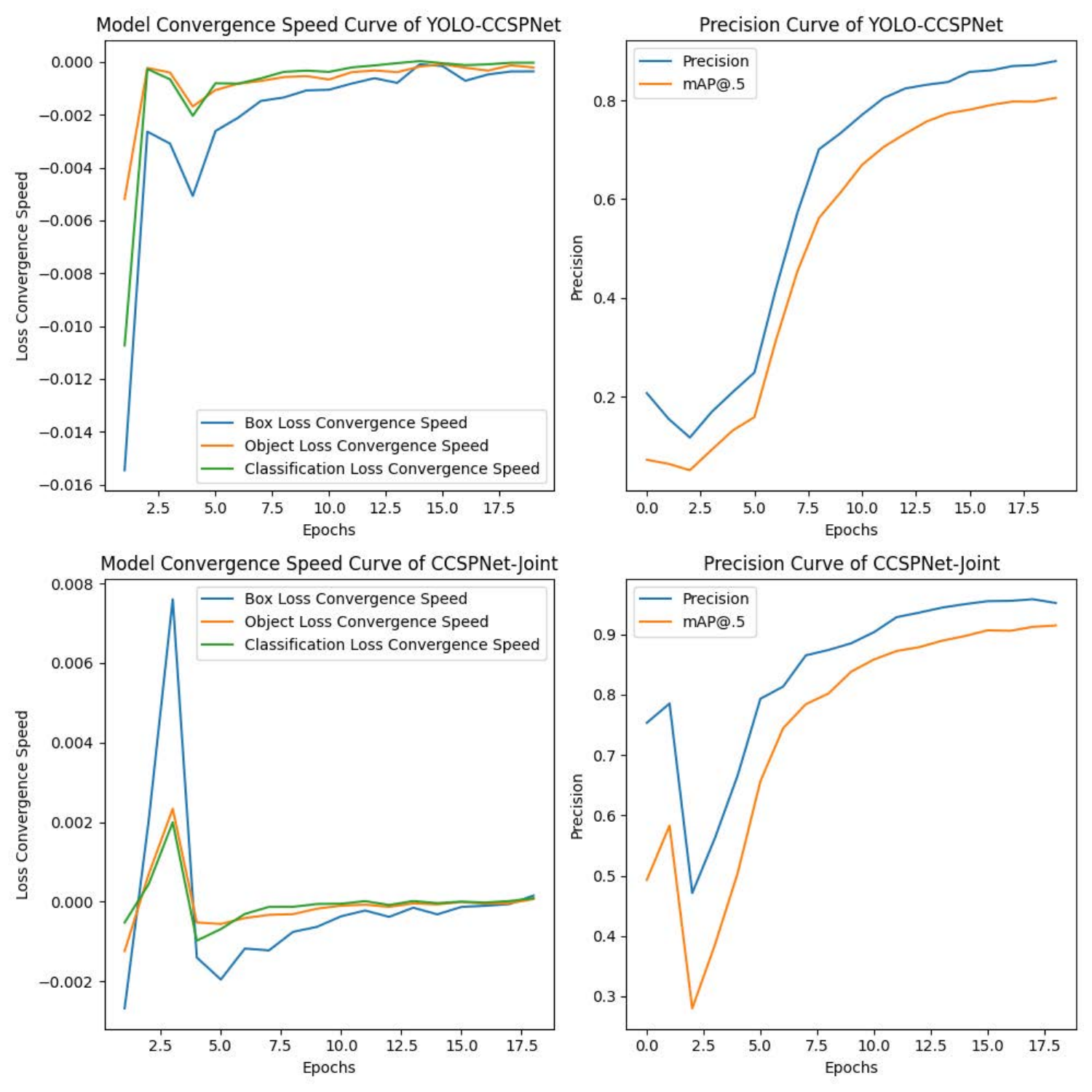}}
\caption{The convergence speed and accuracy comparison of YOLO-CCSPNet and CCSPNet-Joint.}
\label{fig:res}
\end{figure}

\begin{figure*}[h]
  \centering
  \centerline{\includegraphics[width=16cm]{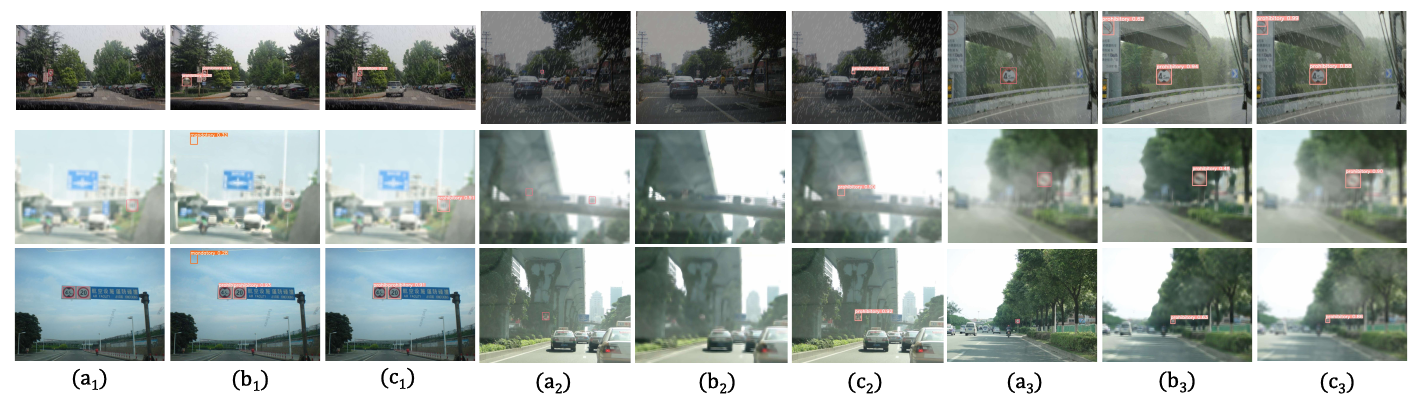}}
  \caption{The visualization of the experimental results is displayed from top to bottom for conditions of rainy weather, foggy weather, and motion blur. (a) Ground Truth, (b) Detection results of the end-to-end training method, and (c) Detection results of the joint training method.}
  \label{fig:res}
\end{figure*}

There are two common training methods for traffic sign detection models applicable to extreme conditions. The first method involves training the object detection model directly on datasets with extreme conditions. The second method involves training an image-denoising model on a dataset with extreme conditions and then training the object detection model on the original dataset in conjunction with the end-to-end use of the image-denoising model. In this study, we propose YOLO-CCSPNet and combine it with the 4kDehazing denoising model for joint training. The experimental results are shown in \textbf{Table \uppercase\expandafter{\romannumeral2}}. It can be observed that through joint training, the model's detection accuracy in extreme weather conditions can be further improved. The final mAP@.5 reaches 0.914, which is a 5.3\% improvement compared to the direct training method and an 18.9\% improvement compared to the end-to-end training method. The mAP@.75 reaches 0.578, which is an 11.58\% improvement compared to the direct training method and a 23.77\% improvement compared to the end-to-end training method. Furthermore, applying the same joint training strategy to the baseline model leads to significant accuracy improvements compared to the other two methods, indicating a certain level of generalization in this approach.\par

The utilization of YOLO-CCSPNet for direct training and the implementation of the joint training approach, CCSPNet-Joint, were compared in terms of the convergence curves of the loss function and the accuracy variations during the model training process, as depicted in \textbf{Fig. 5}. It is evident from the results that the joint-training method exhibits a faster convergence rate of the loss function at the same epoch, and it achieves significantly higher accuracy compared to the direct-training method.\par

\subsection{Ablation Study}
To further demonstrate the effectiveness of our proposed module, we conducted ablation experiments. The experimental results are presented in \textbf{Table \uppercase\expandafter{\romannumeral3}}. It can be observed that replacing the backbone network of the baseline with EfficientViT significantly improves the model's detection accuracy, with a 1.43\% increase in map@.5 and a 1.18\% increase in map@.75. Building upon EfficientViT as the backbone network, we further introduce our module CCSPNet. As a result, the map@.5 reaches 0.861, a 2.86\% improvement over the baseline, and the map@.75 reaches 0.518, a 2.57\% improvement over the baseline.\par

\begin{table}[h]
\centering
\caption{Ablation results in CCSTDB-AUG}
\resizebox{8.5cm}{!}
{
\begin{tblr}{
  row{1} = {c},
  cell{2}{2} = {c},
  cell{2}{3} = {c},
  cell{2}{4} = {c},
  cell{2}{5} = {c},
  cell{3}{2} = {c},
  cell{3}{3} = {c},
  cell{3}{4} = {c},
  cell{3}{5} = {c},
  cell{4}{2} = {c},
  cell{4}{3} = {c},
  cell{4}{4} = {c},
  cell{4}{5} = {c},
  hline{1,5} = {-}{0.08em},
  hline{2} = {-}{},
}
Model                         & Precision & Recall & map@.5 & map@.75 \\
YOLOv5l\textbf{(baseline)}                      & 0.914     & 0.752  & 0.838  & 0.511   \\
YOLOv5l+EfficientViT         & \textbf{0.921}     & 0.762  & \textcolor{red}{0.849} $\uparrow$ \textcolor{cyan}{+0.011}  & \textcolor{red}{0.514} $\uparrow$ \textcolor{cyan}{+0.003}   \\
YOLOv5l+EfficientViT+CCSPNet\textbf{(YOLO-CCSPNet)} & 0.917     & \textbf{0.774}  & \textcolor{red}{0.861} $\uparrow$ \textcolor{cyan}{+0.012}  & \textcolor{red}{0.518} $\uparrow$ \textcolor{cyan}{+ 0.004}   
\end{tblr}
}
\end{table}

\subsection{Visualization}
We conducted a visual analysis of the experimental results, as shown in \textbf{Fig. 6}. From left to right, the sequence is a comparison of false detection scenarios, a comparison of missed detection scenarios, and an improvement in the detection accuracy of the joint training method. We observed that our joint training method, CCSPNet-Joint, achieved higher accuracy in rainy, foggy, and dynamically blurry conditions. Furthermore, under the same detection threshold conditions, CCSPNet-Joint, based on the joint training method, demonstrates fewer missing detections and false detections compared to the end-to-end trained model. This indicates that it meets the safety requirements of intelligent driving.\par

\section{Conclusion}
In this paper, we propose an efficient model, CCSPNet-Joint, based on joint training, which is suitable for traffic sign detection under extreme conditions. Firstly, we design a feature extraction module, CCSPNet, by combining the capturing capability of global features using Transformer and the local feature extraction ability of CNN. By leveraging the contextual information of images, CCSPNet enhances the visual representation capability and effectively improves the feature enhancement ability of the model. Secondly, we utilize EfficientViT as our backbone network in our experiments, which further enhances the model's accuracy. Lastly, we employ joint training by integrating the denoising method 4KDehazing with our model, YOLO-CCSPNet, for mutually dependent training. This further enhances the detection accuracy of the model under extreme conditions.

\begin{refcontext}[sorting = none]
\printbibliography

@article{faisal2019understanding,
  title={Understanding autonomous vehicles},
  author={Faisal, A. and Kamruzzaman, M. and Yigitcanlar, T. and others},
  journal={J. Transp. Land Use},
  volume={12},
  number={1},
  pages={45--72},
  year={2019},
  publisher={JSTOR}
}

@article{yang2015towards,
  title={Towards real-time traffic sign detection and classification},
  author={Yang, Y. and Luo, H. and Xu, H. and others},
  journal={IEEE trans Intell Transp Syst},
  volume={17},
  number={7},
  pages={2022--2031},
  year={2015},
  publisher={IEEE}
}

@article{saadna2017overview,
  title={An overview of traffic sign detection and classification methods},
  author={Saadna, Y. and Behloul, A.},
  journal={IJMIR},
  volume={6},
  pages={193--210},
  year={2017},
  publisher={Springer}
}

@article{maini2009study,
  title={Study and comparison of various image edge detection techniques},
  author={Maini, R. and Aggarwal, H.},
  journal={IJIP},
  volume={3},
  number={1},
  pages={1--11},
  year={2009}
}

@article{he2012guided,
  title={Guided image filtering},
  author={He, K. and Sun, J. and Tang, X.},
  journal={TPAMI},
  volume={35},
  number={6},
  pages={1397--1409},
  year={2012},
  publisher={IEEE}
}

@book{dougherty2003hands,
  title={Hands-on morphological image processing},
  author={Dougherty, E. R and Lotufo, R. A},
  volume={59},
  year={2003},
  publisher={SPIE press}
}

@article{lecun1998gradient,
  title={Gradient-based learning applied to document recognition},
  author={LeCun, Y. and Bottou, L. and Bengio, Y. and others},
  journal={Proc IEEE},
  volume={86},
  number={11},
  pages={2278--2324},
  year={1998},
  publisher={Ieee}
}

@inproceedings{krizhevsky2012imagenet,
  title={Imagenet classification with deep convolutional neural networks},
  author={Krizhevsky, A. and Sutskever, I. and Hinton, G. E},
  booktitle={NeurIPS},
  volume={25},
  year={2012}
}

@inproceedings{vaswani2017attention,
  title={Attention is all you need},
  author={Vaswani, A. and Shazeer, N. and Parmar, N. and others},
  booktitle={NeurIPS},
  volume={30},
  year={2017}
}

@inproceedings{dosovitskiy2020image,
  title={An image is worth 16x16 words: Transformers for image recognition at scale},
  author={Dosovitskiy, A. and Beyer, L. and Kolesnikov, A. and others},
  booktitle={ICLR},
  year={2021}
}

@inproceedings{wang2021pyramid,
  title={Pyramid vision transformer: A versatile backbone for dense prediction without convolutions},
  author={Wang, W. and Xie, E. and Li, X. and others},
  booktitle={CVPR},
  pages={568--578},
  year={2021}
}

@inproceedings{carion2020end,
  title={End-to-end object detection with transformers},
  author={Carion, N. and Massa, F. and Synnaeve, G. and others},
  booktitle={ECCV},
  pages={213--229},
  year={2020},
}

@inproceedings{liu2021swin,
  title={Swin transformer: Hierarchical vision transformer using shifted windows},
  author={Liu, Z. and Lin, Y. and Cao, Y. and others},
  booktitle={ICCV},
  pages={10012--10022},
  year={2021}
}

@inproceedings{liu2022swin,
  title={Swin transformer v2: Scaling up capacity and resolution},
  author={Liu, Z. and Hu, H. and Lin, Y. and others},
  booktitle={CVPR},
  pages={12009--12019},
  year={2022}
}

@article{zhang2020lightweight,
  title={Lightweight deep network for traffic sign classification},
  author={Zhang, J. and Wang, W. and Lu, C. and others},
  journal={ANN TELECOMMUN},
  volume={75},
  pages={369--379},
  year={2020},
  publisher={Springer}
}

@article{zhang2017real,
  title={A real-time Chinese traffic sign detection algorithm based on modified YOLOv2},
  author={Zhang, J. and Huang, M. and Jin, X. and others},
  journal={Algorithms},
  volume={10},
  number={4},
  pages={127},
  year={2017},
  publisher={MDPI}
}

@inproceedings{ren2015faster,
  title={Faster r-cnn: Towards real-time object detection with region proposal networks},
  author={Ren, S. and He, K. and Girshick, R. and others},
  booktitle={NeurIPS},
  volume={28},
  year={2015}
}

@inproceedings{redmon2016you,
  title={You only look once: Unified, real-time object detection},
  author={Redmon, J. and Divvala, S. and Girshick, R. and others},
  booktitle={CVPR},
  pages={779--788},
  year={2016}
}

@article{bochkovskiy2020yolov4,
  title={Yolov4: Optimal speed and accuracy of object detection},
  author={Bochkovskiy, A. and Wang, C. and Liao, H. Mark},
  journal={arXiv:2004.10934},
  year={2020}
}

@article{jocher2022ultralytics,
  title={ultralytics/yolov5: v7. 0-yolov5 sota realtime instance segmentation},
  author={Jocher, G. and Chaurasia, A. and Stoken, A. and others},
  journal={Zenodo},
  year={2022}
}

@inproceedings{wang2023yolov7,
  title={YOLOv7: Trainable bag-of-freebies sets new state-of-the-art for real-time object detectors},
  author={Wang, C. and Bochkovskiy, A. and Liao, H. Mark},
  booktitle={CVPR},
  pages={7464--7475},
  year={2023}
}

@inproceedings{yasarla2020syn2real,
  title={Syn2real transfer learning for image deraining using gaussian processes},
  author={Yasarla, R. and Sindagi, V. A and Patel, V. M},
  booktitle={CVPR},
  pages={2726--2736},
  year={2020}
}

@inproceedings{guo2021efficientderain,
  title={Efficientderain: Learning pixel-wise dilation filtering for high-efficiency single-image deraining},
  author={Guo, Q. and Sun, J. and Xu, J. F and others},
  booktitle={AAAI},
  volume={35},
  number={2},
  pages={1487--1495},
  year={2021}
}

@article{he2010single,
  title={Single image haze removal using dark channel prior},
  author={He, K. and Sun, J. and Tang, X.},
  journal={TPAMI},
  volume={33},
  number={12},
  pages={2341--2353},
  year={2010},
  publisher={IEEE}
}

@inproceedings{zheng2021ultra,
  title={Ultra-high-definition image dehazing via multi-guided bilateral learning},
  author={Zheng, Z. and Ren, W. and Cao, X. and others},
  booktitle={CVPR},
  pages={16180--16189},
  year={2021},
}

@inproceedings{tsai2022stripformer,
  title={Stripformer: Strip transformer for fast image deblurring},
  author={Tsai, F. and Peng, Y. and Lin, Y. and others},
  booktitle={ECCV},
  pages={146--162},
  year={2022},
}

@article{li2022contextual,
  title={Contextual transformer networks for visual recognition},
  author={Li, Y. and Yao, T. and Pan, Y. and others},
  journal={TPAMI},
  volume={45},
  number={2},
  pages={1489--1500},
  year={2022},
  publisher={IEEE}
}

@inproceedings{wang2020cspnet,
  title={CSPNet: A new backbone that can enhance learning capability of CNN},
  author={Wang, C. and Liao, H. M and Wu, Y. and others},
  booktitle={CVPR workshops},
  pages={390--391},
  year={2020}
}

@inproceedings{he2017mask,
  title={Mask r-cnn},
  author={He, K. and Gkioxari, G. and Doll{\'a}r, P. and others},
  booktitle={ICCV},
  pages={2961--2969},
  year={2017}
}

@article{terven2023comprehensive,
  title={A comprehensive review of YOLO: From YOLOv1 to YOLOv8 and beyond},
  author={Terven, J. and Cordova-Esparza, D.},
  journal={arXiv:2304.00501},
  year={2023}
}

@article{lu2018traffic,
  title={Traffic signal detection and classification in street views using an attention model},
  author={Lu, Y. and Lu, J. and Zhang, S. and others},
  journal={Comput Vis Media},
  volume={4},
  pages={253--266},
  year={2018},
  publisher={Springer}
}

@article{dong2021accurate,
  title={An accurate small signal dynamic model for LCC-HVDC},
  author={Dong, Y. and Ma, J. and Wang, S. and others},
  journal={IEEE Trans. Appl. Supercond},
  volume={31},
  number={8},
  pages={1--6},
  year={2021},
  publisher={IEEE}
}

@article{hou2022traffic,
  title={Traffic signs detection and recognition systems by light-weight multi-stage network},
  author={Hou, M. and Zhang, X. and Chen, Y. and others},
  journal={Multimed Tools Appl},
  volume={81},
  number={12},
  pages={16155--16169},
  year={2022},
  publisher={Springer}
}

@inproceedings{chen2019hybrid,
  title={Hybrid task cascade for instance segmentation},
  author={Chen, K. and Pang, J. and Wang, J. and others},
  booktitle={CVPR},
  pages={4974--4983},
  year={2019}
}

@inproceedings{xu2023multi,
  title={Multi-Task Learning with Knowledge Distillation for Dense Prediction},
  author={Xu, Y. and Yang, Y. and Zhang, L.},
  booktitle={CVPR},
  pages={21550--21559},
  year={2023}
}

@inproceedings{liu2023efficientvit,
  title={EfficientViT: Memory Efficient Vision Transformer with Cascaded Group Attention},
  author={Liu, X. and Peng, H. and Zheng, N. and others},
  booktitle={CVPR},
  pages={14420--14430},
  year={2023}
}
\end{refcontext}

\end{document}